# A Hybrid Framework for Tumor Saliency Estimation

Fei Xu[1], Min Xian[2], Yingtao Zhang[3], Kuan Huang[2], H. D. Cheng[1,3], Boyu Zhang[1],
Jianrui Ding[4], Chunping Ning[5], Ying Wang[6*]

[1] Department of Computer Science, Utah State University, Logan, USA
[2] Department of Computer Science, University of Idaho, Idaho Falls, USA
[3] School of Computer Science and Technology, Harbin Institute of Technology, Harbin, China
[4] School of Computer Science and Technology, Harbin Institute of Technology, Weihai, China
[5] Department of Ultrasound, The Affiliated Hospital of Qingdao University, Qingdao, China
[6] Department of General Surgery, The Second Hospital of Hebei Medical University, Shijiazhuang, China

*Abstract*—Automatic tumor segmentation of breast ultrasound (BUS) image is quite challenging due to the complicated anatomic structure of breast and poor image quality. Most tumor segmentation approaches achieve good performance on BUS images collected in controlled settings; however, the performance degrades greatly with BUS images from different sources. Tumor saliency estimation (TSE) has attracted increasing attention to solve the problem by modeling radiologists' attention mechanism. In this paper, we propose a novel hybrid framework for TSE, which integrates both high-level domain-knowledge and robust low-level saliency assumptions and can overcome drawbacks caused by direct mapping in traditional TSE approaches. The new framework integrated the Neutro-Connectedness (NC) map, the adaptive-center, the correlation and the layer structure-based weighted map. The experimental results demonstrate that the proposed approach outperforms state-of-the-art TSE methods.

## I. Introduction

Breast cancer is the most frequently diagnosed cancer and account for about 29% of all new female cancer cases [1]. Automatic BUS segmentation is a key component in computer-aided diagnosis (CAD) systems and has the advantages of operator-independence and high reproducibility [2, 3]. However, developing automatic segmentation approaches for BUS images is challenging due to the speckle noise, low contrast, weak boundary, and artifacts; furthermore, strong priors to object features such as tumor size, shape and echo strength vary considerably across patients and machine settings, and cannot work well on images from multiple sources [19].

Many automatic BUS segmentation approaches have been proposed in the last decade [3,18, 21, 33-36]. The major strategy of the approaches is to locate tumors automatically by modeling domain-related priors. However, some strong constraints such as the number of tumors, tumor size, and predefined tumor locations, were utilized in the approaches, which result in dramatic performance degradation in clinical practice where BUS images could be collected under different settings or situations such as low image contrast, more artifacts, containing no tumor/more than one tumors per image, etc. Therefore, it is crucial to develop automatic BUS segmentation techniques that are invariant and robust to images settings.

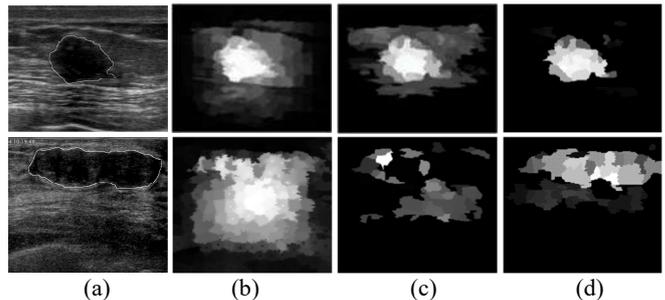

Fig. 1. Tumor saliency estimation examples. (a)Two BUS images with the ground truths (white boundaries); (b) results of the method in [17]; (c) results of the method in [21]; (d) results of the proposed method.

Visual saliency estimation (VSE) is an important and popular method to achieve the unconstrained image segmentation [20-22] by modeling human visual mechanism. It measures the degrees of different image regions attracting human attention. The center, and the local and global contrast are typical clues modeled in VSE approaches. Generally, VSE approaches can be classified into two categories based on the ways of saliency generation. First, the directly mapping methods [4-9] transfer image features into saliency values using predefined mapping; second, the optimization models [10-17] focus on modeling different hypotheses into one framework, and the saliency values are generated by optimization techniques. The approaches in the former category are faster; however, they have lower accuracy because the directly mapping methods fail on images with low contrast and big objects. While approaches in the latter category can achieve better performance by automatically adapting and balancing different elements of the models. Most VSE models were proposed to process natural images utilizing bottom-up frameworks and cannot achieve good performance on BUS images (Fig.1).

TSE aims to model the visual clues of tumors that attract radiologists' attention during the tumor segmentation. The TSE outputs the saliency value of each BUS image pixel in terms of the pixel's possibility of belonging to a tumor. In [21], Shao et al. proposed a TSE model for fully automatic tumor segmentation. The model combined tumor prior knowledge and saliency estimation hypothesis and achieved very good performance using their BUS image dataset. However, it has two main drawbacks: 1) it always outputs a salient region and cannot deal with images without tumor; 2) the predefined mapping failed to handle the images with large tumors, shadows, and low contrast (Fig. 1). Xie et al. [20] proposed to calculate TSE by modeling intensity, blackness ratio, and superpixel contrast; and the final saliency value of each pixel was the average of values of the three components. It shares the same drawbacks with [20] due to the nature of direct mapping and the strategy of "winner-take-all".

*This work is supported, in part, by the Chinese NSF (81501477) and by the Livelihood Technology Project of Qingdao city (15-9-2-89-NSH)

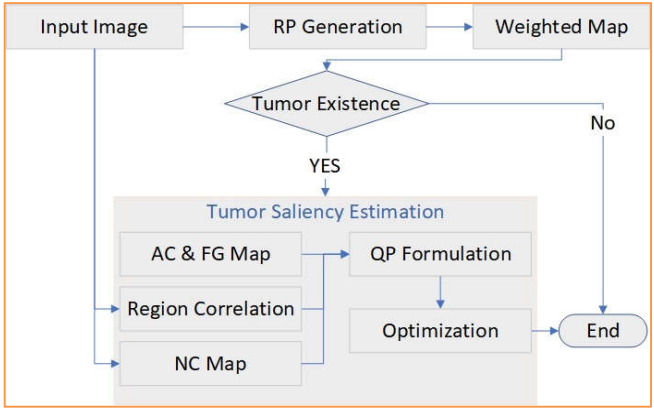

Fig. 2. Flowchart of the proposed method.

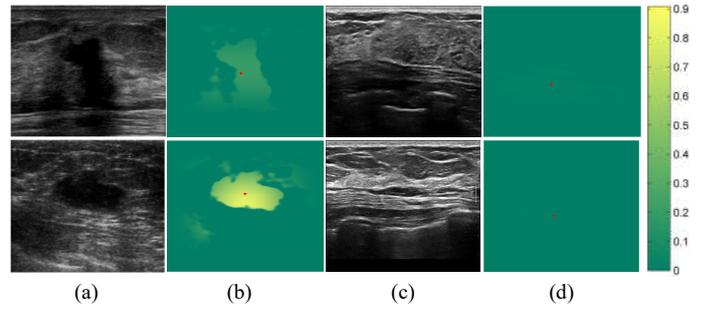

Fig. 3. Weighted maps. (a) BUS images with tumor; (b) and (d) weighted maps with RPs (marked with green color); (c) BUS images without tumor.

Xu et al. proposed a general bottom-up saliency estimation model [22] that integrated many robust hypotheses: the global contrast, adaptive center-bias, boundary constraint and the smoothness term based on the color statistic. The model is flexible, and the global optimum can be reached by using the primal-dual interior point method. However, the model always outputs a salient region and cannot handle BUS images without tumors.

To solve the problems, we propose a novel hybrid framework for TSE, which follows a two-step strategy. The first step determines if a BUS image has tumor (s) based on the weighted map by utilizing adaptive reference point (RP) generation [3]. In the second step, it formulates the TSE problem as a quadratic programming (QP) problem which integrates both high-level domain-knowledge and robust low-level saliency assumptions. In this framework, it incorporates Neutro-Connectedness (NC) [31] to generate more robust and accurate boundary connectedness, and to measure the corresponding degree of confidence simultaneously. The adaptive center-bias (AC), foreground map (FG Map), which is based on the weighted map, and regions' correlation hypothesis are also integrated into the framework. The flowchart of the proposed model is shown in Fig.2.

## II. PROPOSED METHOD

### A. Tumor existence determination

Existing TSE methods assume that there exists a tumor in each BUS image and cannot handle the image without tumor; however, as an automatic tumor detection or segmentation system, it is important to identify whether there is a tumor or not. Besides, the convex optimization frameworks [22] cannot deal with the image without salient object. The equality constraint will force that there must be at least one salient object in an image.

In [3], Xian et al. proposed an algorithm to automatically generate the adaptive reference point (RP) based on the breast anatomy. The RP is generated accurately and fast and can detect the darker regions (candidates of the tumors). The weighted map is constructed based on the RPs and the intensity map. The region is nearer the RP, the intensity value of the region in the weighted map is higher, vice versa. As shown in Fig.3, the weighted map will enhance the low-intensity pixels near RP and decay the high-intensity pixels far away from RP.

Based on the observation, the weighted map of the BUS image without tumor is smoother than that of the BUS images with tumor. In Fig. 3, the four max intensities of the weighted maps are 0.043, 0.0152, 0.9086 and 0.0035, listing from left to right, top to bottom respectively. It chooses the local maximum, mean, and standard deviation of the weighted map as the feature vector and applies the threshold or Decision Tree. The result and discussion are in Section III.

### B. Tumor Saliency Estimation

Researchers have applied several saliency hypotheses to construct mathematics models for VSE, such as rarity hypothesis, center-bias hypothesis, correlation hypothesis, etc. In this work, it utilizes the adaptive center-bias, regions' correlation hypotheses, the boundary NC and weighted maps to model the TSE problem as a convex optimization problem.

Firstly, it used a quick shift algorithm in [24] to over-segment the image into $N$ superpixels, noted as $\{R_i\}_{i=1}^{i=N}$. Similar to the method in [21], it extracts regions' average intensities as the region features. To facilitate the discussion, it defines $S = (s_1, s_2, \cdots, s_n)^T$ as a vector of saliency values for $N$ image regions, where $s_i$ denotes the saliency value of the $i$th image region and $s_i \in [0,1]$. The optimization of the model is to assign the optimal saliency values for a set of image regions.

#### 1) Problem formulation

The problem is formulated as

$$\text{minimize } E(S) = S^T(T + \alpha C + \gamma W) +$$
$$+ \beta \sum_{i=1}^{N}\sum_{j=1}^{N}(s_i - s_j)^2 r_{ij} D_{ij} \quad (1)$$
$$\text{subject to } 0 \leq s_i \leq 1, i = 1,2,\cdots,N;$$
$$\sum_{i=1}^{N} s_i = 1$$

In Eq. (1), the term $T = (t_1, t_2, \cdots, t_N)^T$ denotes the NC map, and $t_i$ defines the NC between the $i$th region and the boundary; the term $C = (c_1, c_2, \cdots, c_N)^T$ is the distance map, and $c_i$ defines the distance between the $i$th region and the adaptive-center; the term $W = (w_1, w_2, \cdots, w_N)^T$ is the FG map, and $w_i$ is the value of the $i$th region; the terms $r_{ij}$ and $D_{ij}$ define the similarity and the spatial distance between the $i$th and the $j$th regions, respectively. The term $S^T T$ defines the cost using the NC map, the term $S^T W$ defines the cost on the FG map, and the term $S^T C$ defines the cost based on the adaptive center-bias. The last term is the smoothness that forces the regions with similar features to have similar saliency values.

The formulated problem is a typical QP problem with linear equality and inequality constraints. The original problem can be rewritten as follows:
$$\text{minimize } f_0(S) = \left(\sum_{i=1}^{N} s_i t_i + \alpha \sum_{i=1}^{N} s_i c_i + \gamma \sum_{i=1}^{N} s_i w_i\right)$$
$$+ \beta \sum_{i=1}^{N}\sum_{j=1}^{N}(s_i - s_j)^2 r_{ij} D_{ij} \quad (2)$$
$$\text{subject to } 0 \leq s_i \leq 1, i = 1,2,\cdots,N;$$
$$\sum_{i=1}^{N} s_i = 1$$

*2) NC map generating*

Boundary connectivity is an effective prior utilized in many visual saliency estimation models [22, 26-30]. Most models define the boundary connectivity by using the shortest path between the local regions and the boundary. However, such connectivity cannot handle noisy data well. The Neutro-Connectedness (NC) theory [31, 32] introduces a new domain, the degree of confidence, to measure the confidence of the connectedness. The new domain is very useful to avoid the fake connectedness caused by the uncertainty, such as noise.

In [32], the NC of two region contains three parts: the degree of truth, the degree of confidence, and the degree of false, $NC(i,j) = [T(i,j), I(i,j), 1 - T(i,j)]$ where $i$ and $j$ indicate the $i$th and $j$th pixel or region, respectively.

Here, NC map is defined on image region $\{R_i\}_{i=1}^{i=N}$. To calculate the NC triplet between all the regions with the boundary set, it applies the definitions of NC and computation algorithm in [31, 32]. For more details of NC theory, refer [31, 32]. The three basic ideas of NC are summarized as follows:

*NC of two adjacent regions $i$ and $j$*

$$\mu_T(i,j) = exp(-|Gray_i - Gray_j|/\sigma^2) \quad (3)$$
$$\mu_I(i,j) = 1 - max(h(i), h(j)) \quad (4)$$

where $Gray_i$ is the regions average gray level of the $i$th region, $\sigma^2 = 0.5$, and $h(i)$ is the inhomogeneity of the $i$th region [31, 32].

*NC of a path.* The degree connectedness of a path is defined as the minimum value of $\mu_T$ along the path, and the confidence is the maximum $\mu_I$ value along the path.

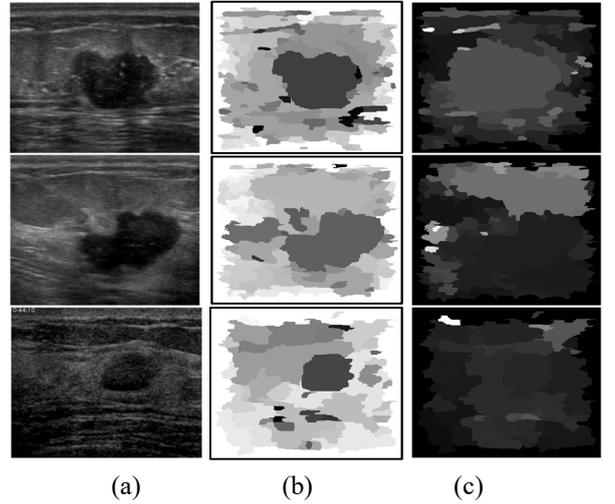

Fig. 4. $T$ map and $I$ map samples. (a) BUS images with tumor; (b) $T$ maps; (c) $I$ maps.

*NC of any two regions.* The degree of connectedness is defined by the strongest path of all paths connecting the two regions. It uses the confidence of the strongest path as the degree of confidence of the two regions. For more information about how to deal with ties, refer [31, 32].

As the particular characteristic that no tumor is touching the border, it sets the border regions as the background seeds to generate the NC map by using the algorithm in [31]. Fig. 4 shows some samples, $T$ maps, which refers to the background map generated by the NC theory, and $I$ maps, which refers to the confidence of the degree of NC between each local region and boundary set. Based on the three basic ideas discussed, the $I$ map will be used to generate each path of NC map.

$T$ and $I$ maps of image regions employed in the framework are defined as $T = (t_1, t_2, \cdots, t_N)^T$ and $I = (I_1, I_2, \cdots, I_N)^T$, and they are $N \times 1$ vectors.

*3) Adaptive center bias and FG maps generation*

Traditional saliency estimation models usually use the image center as an important visual clue to estimate the saliency map. However, it fails when objects are far away from the center. The approach in [22] solve this problem on natural images by estimating the adaptive center using the weighted local contrast map; but the local contrast map is sensitive to noise and cannot achieve good performance on BUS images. Instead of detecting the top and bottom lines of mammary layer [21], a new tumor detection approach was proposed by utilizing the RP and weighted map [31].

The FG map in the framework is defined as $W = (w_1, w_2, \cdots, w_N)^T$

$$w_i = exp(-m_i/\sigma^2) \quad (5)$$

where $m_i$ is the mean value of the $i$th region in the weighted map, and $\sigma^2 = 0.5$.

The reference point is used as the adaptive center in the saliency detection model. It is defined as $C = (c_1, c_2, \cdots, c_N)^T$

$$c_i = exp(\|SC_i - RP\|_2/d_D) \quad (6)$$

where $SC_i$ is the coordinate of the $i$th region's center and the value is in [0,1]. $RP$ is the reference point position. $\|\cdot\|_2$ is the $l_2$ norm. $d_D$ is equal to $\sqrt{2}/2$.

*4) Regions' correlation*

It uses the region correlation hypothesis to force the closer similar regions to have similar saliency value.

It defines $w_{ij}$ as the similarity, and $D_{ij}$ as the spatial distance between the $i$th and the $j$th regions.

$$w_{ij} = \exp(-|Gray_i - Gray_j|/\sigma^2) \quad (7)$$
$$D_{ij} = \exp(-\|SC_i - SC_j\|_2/d_D) \quad (8)$$

where *Gray* is the regions average gray level vector, In Eq. (8), $SC_i$ is the coordinate of the $i$th region's center and the value is in [0,1]. $\sigma^2 = 0.5$. $d_D$ is equal to $\sqrt{2}/2$. $\|\cdot\|_2$ is the $l_2$ norm.

## C. Optimization

It uses the primal-dual method to optimize the QP with linear inequality and equality constraints similar with [22]. It can obtain the global optimal value quickly. There are three important steps to apply the primal-dual interior point method: (1) modify the KKT conditions and obtain the dual, prime and centrality residuals; (2) obtain the primal-dual search direction; and (3) update $S$ and the dual variables.

In the primal-dual interior method, $t^0$ and $S^0$ are initialized as 1 and $(1/N)E^T$, respectively; and the dual residual, primal residual, and the centrality residual are updated in each iteration, and the optimization processing stops when the sum of the $l_2$ norms is less than $10^{-6}$.

## III. EXPERIMENTAL RESULTS

### A. Datasets, metrics and setting

In this section, it validates the performance of the newly proposed method using a BUS image dataset containing 706 ultrasound images from heterogeneous sources, in which 96 images have no tumors, and 610 images have tumors [25]. All experiments are performed by using Matlab (R2014a, MathWorks Inc., MA) on a Windows-based PC equipped with a dual-core (3.6 GHz) processor and 8 GB memory.

**Metrics of saliency estimation**: the saliency estimation is evaluated using the 610 images with tumors. The precision-recall (P-R) curve, F-measure and mean absolute error (MAE) are employed to evaluate the overall performance of saliency detection method. The precision and recall ratios are defined as follows:

$$Precison = \frac{|SM \cap GT|}{|SM|}, Recall = \frac{|SM \cap GT|}{|GT|}$$

where *SM* denotes the binary saliency map, while *GT* is the ground truth binary map, and $|SM|$ denotes the white pixel number of the saliency map. The P-R curve shows the mean precision and recall rate of all saliency maps on a dataset. For each method, the P-R curve is calculated by segmenting saliency map with threshold range from 0 to 255, and computing the precision and recall rates by comparing the thresholding result with the ground truth. To obtain the average precision and recall rates, it uses an adaptive thresholding method [8], which chooses two times the mean saliency value as the threshold. The F-measure [9] and MAE [26] are defined as

$$F_\gamma = \frac{(1+\gamma^2)Precison \cdot Recall}{\gamma^2 \cdot Precison + Recall}$$
$$MAE = \sum_{i=1}^{M} |S(p_i) - G(p_i)|$$

where $\gamma^2$ is set to 0.3 as suggested in published saliency detection methods, $p_i$ is the coordinate of the $i$th image pixel, $S(p_i)$ is the saliency value of the $i$th pixel, and $G$ is the binary ground truth.

**Metrics of tumor existence determination**: two metrics, precision ratio, recall ratio are utilized:

$$PR = \frac{|TC|}{|TS|}, RR = \frac{|TC|}{|Tmset|}$$

where $|TC|$ is the number of correct detected images with tumors, and $|TS|$ is the total number of images detected with tumors; $|Tmset|$ is the total number of images with tumors in the dataset.

**Parameter setting:** all the experiments are based on the parameters: $\alpha = 10, \beta = 2$, and $\gamma = 80$.

### B. Tumor existence determination

Based on the observation, the maximum value of the weighted map is very useful to identify whether there is a tumor. Simple thresholds were applied to the maximum value, and the result is shown in Table 1. There are no distinguish differences (less than 2%) between the values of PR and RR by using different threshold values because of the limitation by the number of images with no tumor. Although the difference is small, it will have a significant impact in the clinical practice. The mean accuracy is 100% if using the Decision Tree classifier with ten-fold cross-validation method.

TABLE 1. Results of theresholding

| Thresholds | PR | RR |
|---|---|---|
| 0.02 | 98.07% | 99.84% |
| 0.03 | 99.84% | 99.51% |
| 0.04 | 99.02% | 99.67% |
| 0.05 | 99.84% | 99.84% |
| 0.057 | 100% | 99.34% |

### C. Tumor saliency estimation evaluation

The proposed method is compared with most recently published methods SMTD [21], OMRC [22], RBD [26] and RRWR [17]. SMTD is the directly mapping method for tumor saliency estimation. And the other three are the bottom-up models and can achieve good performance in the natural images. The proposed method can make the high saliency value concentrate on the tumor and the background areas have low

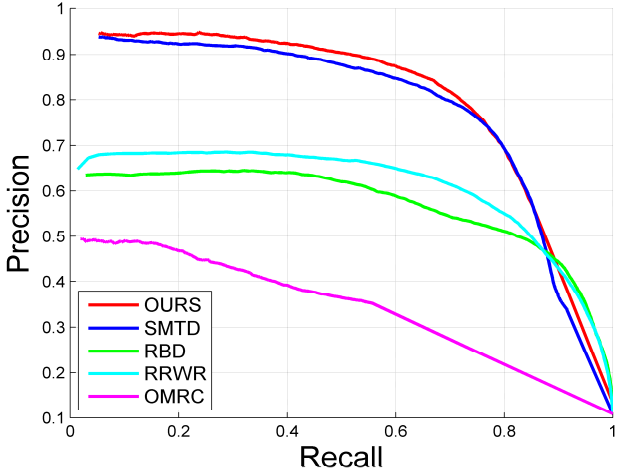

Fig. 5. The PR curve of the five approaches.

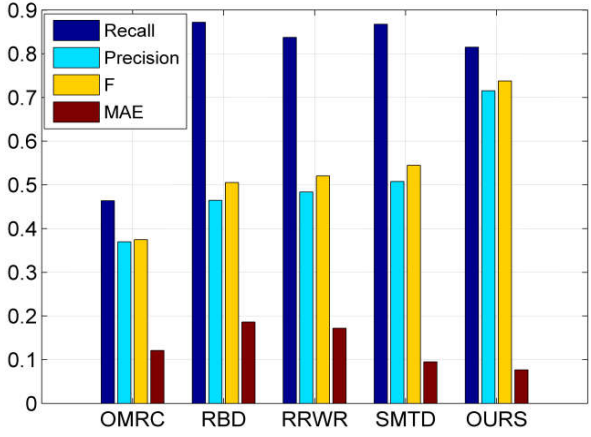

Fig. 6. The recall, precision, F-measure and MAE of five

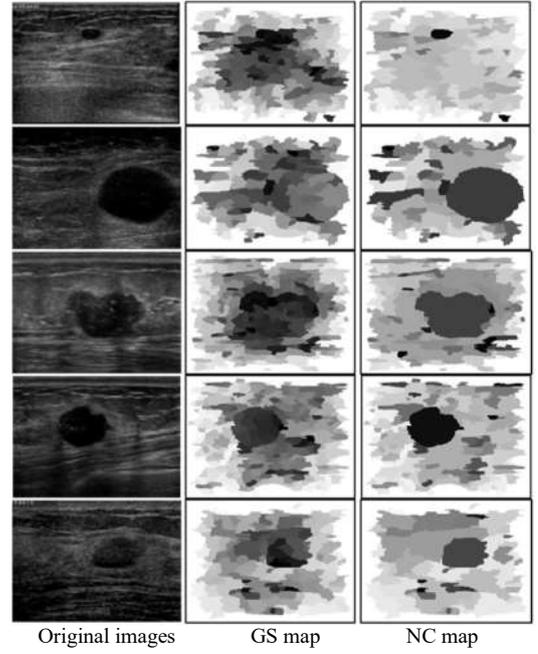

Original images    GS map    NC map

Fig. 7. The GS map and NC map samples.

saliency value while the methods SMTD, RBD and RRWR made the background regions around the tumor have higher salient values. This situation will cause the recall ratio of the methods higher but the precision ratio lower; and the result are more accurate than those of other methods. Especially, SMTD and OMRC cannot produce good saliency maps even miss the big tumors.

The performance of the proposed method is evaluated using the metrics and the dataset: MAE values, the F-measure values, and P-R curves. As shown in Figs. 5-6, the proposed method is better than other methods. The methods, SMTD, RBD and RRWR, can obtain relative high average recall ratios, but the precision ratios and F-measures are low. It is s because the saliency maps generating by those methods make the tumors as well as the background around the tumors have high saliency value.

### D. The effectiveness of NC term

The NC map with the boundary connectivity based on the graph shortest path is computed. In this experiment, it used the algorithm of [29] and defined the edge weight for each pair of adjacent nodes as $|Gray_i - Gray_j|$ to obtain the background map, where $Gray_i$ is the average gray level of the $i$th region. The examples of 5 images are shown in Fig. 7. The results in

Fig.7 demonstrate that the two methods can both achieve better results on the smooth BUS images (the 3rd and 4th rows). The method based on graph shortest path (GS) fails to handle the BUS images with too small or too large tumors, or poor quality with noise. Moreover, the maps generated by NC method are much smoother than that of GS method.

### E. Parameter tuning

$\alpha$, $\beta$ and $\gamma$. As presented in Section II, the detection framework has 4 major parts. The NC map or weighted map cannot always provide the correct information to generate the saliency map (see Figs 3-4.). It is very important to balance the effect of each part.

The values of $\alpha$, $\beta$ and $\gamma$ are used to balance the influence of the adaptive center-bias term, weighted map and smooth term, respectively. It evaluates the performance of the proposed method with $\alpha$ ranging from 0 to 200, $\beta$ ranging from 0 to 200, and $\gamma$ ranging from 0 to 200 on the randomly selected subset containing 20 images. There are three stages to choose the parameters. The first stage, it makes the step size of three parameters be 40 and roughly obtain the range of each parameter which can achieve better P-R curve performance and MAE value if the P-R curve is similar. The second stage, the step size is 10. The step size is 2 in the third stage. As shown in Fig. 8, we obtain better P-R curve when $\alpha$ is close to 10, $\beta$ is close to 2 and $\gamma$ is close to 80, respectively.

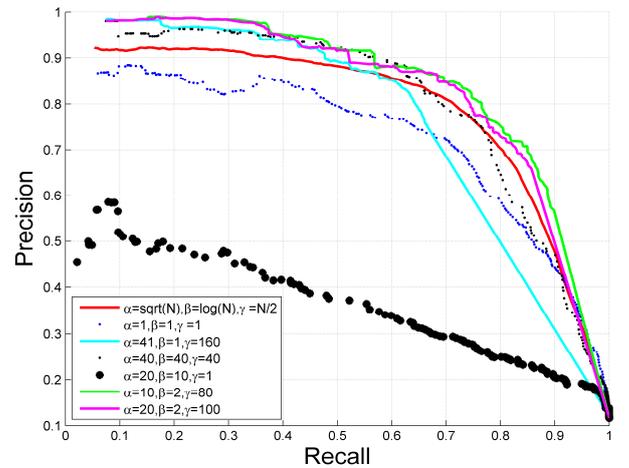

Fig. 8. The P-R curve of different parameter values

## IV. CONCLUSION

In this paper, we propose a hybrid optimization framework for tumor saliency estimation, which models both the domain-related knowledge and low-level visual clue. There are three main reasons that the proposed method can achieve much better results than traditional approaches: (1) the proposed framework integrates the weighted map, NC map and reference point, robust low-level image features; (2) the proposed method model TSE as an optimzation problem which can balance each part of energy function automatically for different images; (3) large amount of BUS images is used to choose the proper parameters.